\documentclass[3p,11pt]{elsarticle}

\usepackage{amssymb}
\usepackage{amsmath}
\usepackage{subfig}
\usepackage{amsmath,amsbsy}
\usepackage{makecell} 
\usepackage{multirow}
\usepackage{hyperref}
\usepackage{xcolor}
\usepackage[normalem]{ulem}
\usepackage{caption}
\usepackage{algorithm}
\DeclareCaptionFormat{myformat}{#3}
\usepackage{color}
\usepackage{listings}

\definecolor{backcolour}{rgb}{0.95,0.95,0.92}
\definecolor{codegreen}{rgb}{0,0.6,0}

\usepackage{natbib}

\usepackage{color, colortbl}
\newcommand{\K}[0]{five }

\definecolor{Gray}{gray}{0.65}
\newcolumntype{g}{>{\columncolor{Gray}}c}

\newcommand{\ME}[1]{\textit{Ours} ($\delta_T=#1$)}
\newcommand{\SH}{Southampton Eye Unit }
\newcommand{\MF}{Moorfields Eye Hospital }

\journal{Medical Image Analysis}

\begin{document}

\begin{frontmatter}

\title{Metadata-enhanced contrastive learning from retinal optical coherence tomography images}

\author[biomedia]{Robbie Holland}
\author[laboia]{Oliver Leingang}
\author[laboia,christian]{Hrvoje Bogunović}
\author[laboia]{Sophie Riedl} 
\author[michigan]{Lars Fritsche}
\author[kcl]{Toby Prevost}
\author[imco_basel,do_basel]{Hendrik P. N. Scholl}
\author[laboia]{Ursula Schmidt-Erfurth}
\author[ucl,moorfields]{Sobha Sivaprasad} 
\author[southampton]{Andrew J. Lotery} 
\author[biomedia]{Daniel Rueckert}
\author[biomedia,tum]{and Martin J. Menten}
\author{on behalf of the PINNACLE consortium}

\affiliation[biomedia]{organization={BioMedIA, Imperial College London},
state={London},
country={United Kingdom}}

\affiliation[laboia]{organization={Laboratory for Ophthalmic Image Analysis, Medical University of Vienna},
state={Vienna},
country={Austria}}

\affiliation[christian]{organization={Christian Doppler Lab for Artificial Intelligence in Retina, Medical University of Vienna},
state={Vienna},
country={Austria}}

\affiliation[michigan]{organization={Department of Biostatistics, University of Michigan},
city={Ann Arbor},
state={MI},
country={United States}}

\affiliation[kcl]{organization={Nightingale-Saunders Clinical Trials \& Epidemiology Unit, King's College London},
city={London},
country={United Kingdom}}

\affiliation[imco_basel]{organization={Institute of Molecular and Clinical Ophthalmology Basel},
city={Basel},
state={Basel-Stadt},
country={Switzerland}}

\affiliation[do_basel]{organization={Department of Ophthalmology, Universitat Basel},
city={Basel},
state={Basel-Stadt},
country={Switzerland}}

\affiliation[ucl]{organization={Institute of Ophthalmology, University College London},
city={London},
country={United Kingdom}}

\affiliation[moorfields]{organization={Moorfields National Institute for Health and Care Biomedical Research Centre, Moorfields Eye Hospital},
city={London},
country={United Kingdom}}

\affiliation[southampton]{organization={Clinical and Experimental Sciences, Faculty of Medicine, University of Southampton},
city={Southampton},
state={Hampshire},
country={United Kingdom}}

\affiliation[tum]{organization={Institute for AI and Informatics in Medicine, Technical University of Munich},
city={Munich},
state={Bavaria},
country={Germany}}

\begin{abstract}

{
Deep learning has potential to automate screening, monitoring and grading of disease in medical images. Pretraining with contrastive learning enables models to extract robust and generalisable features from natural image datasets, facilitating label-efficient downstream image analysis. However, the direct application of conventional contrastive methods to medical datasets introduces two domain-specific issues. Firstly, several image transformations which have been shown to be crucial for effective contrastive learning do not translate from the natural image to the medical image domain. Secondly, the assumption made by conventional methods, that any two images are dissimilar, is systematically misleading in medical datasets depicting the same anatomy and disease. This is exacerbated in longitudinal image datasets that repeatedly image the same patient cohort to monitor their disease progression over time. In this paper we tackle these issues by extending conventional contrastive frameworks with a novel metadata-enhanced strategy. Our approach employs widely available patient metadata to approximate the true set of inter-image contrastive relationships. To this end we employ records for patient identity, eye position (i.e. left or right) and time series information. In experiments using two large longitudinal datasets containing 170,427 retinal optical coherence tomography (OCT) images of 7,912 patients with age-related macular degeneration (AMD), we evaluate the utility of using metadata to incorporate the temporal dynamics of disease progression into pretraining. Our metadata-enhanced approach outperforms both standard contrastive methods and a retinal image foundation model in five out of six image-level downstream tasks related to AMD. We find benefits in both a low-data and high-data regime across tasks ranging from AMD stage and type classification to prediction of visual acuity. Due to its modularity, our method can be quickly and cost-effectively tested to establish the potential benefits of including available metadata in contrastive pretraining.
}
\end{abstract}




\end{frontmatter}


\section{Introduction}
Deep learning promises to revolutionise medical imaging by automating screening, diagnosis and monitoring of diseases \citep{davenport2019potential}. Traditionally, the successful application of supervised deep learning has required large amounts of labelled data for training. This does not translate well to the medical domain where labelled images are scarce as their annotation typically requires the time and cost of a trained clinician. ImageNet, commonly used to train and benchmark algorithms for natural image processing, contains millions of labelled images \citep{deng2009imagenet}, while the median size of datasets featured in the Medical Image Computing \& Computer Assisted Intervention (MICCAI) conference in 2019 ranged from 120 to 180 subjects \citep{kiryati2021dataset}. There is a large disparity between the growing amount of medical data and the resources needed to annotate it for the research and development of deep learning tools in medicine \citep{willemink2020preparing}.\\
Pretraining with contrastive frameworks enables deep learning models to reach the same level of performance using fewer labelled training samples \citep{chen2020big}. They work by contrasting two augmented views of each image per batch, assuming their similarity depends on whether or not they originate from the same image. However, they are prone to provide misleading learning signals when the true set of inter-image relationships are unknown. {As illustrated in \citep{khosla2020supervised}, removing negative pairs of images from the same ImageNet class improves downstream performance. In contrast, large volumes of medical images are not typically annotated with gold-standard class labels for disease stage or type, but are commonly accompanied by metadata such as the date of the medical scan and the patient's anonymised identity.} This widely available information has the potential to indicate the true set of inter-image relationships used in contrastive frameworks in the medical domain.\\
Optical coherence tomography (OCT) enables low-cost, non-invasive imaging of the eye. This has led to the accumulation of large, metadata-enriched retinal datasets that approach the size of those for natural images. Hence, ophthalmology is ideally suited for adapting advances in self-supervised learning to the medical domain. A particularly challenging task is the diagnosis and prognosis of age-related macular degeneration (AMD) in retinal OCT. AMD, the leading cause of irreversible blindness in the elderly, is a progressive disease and is that is projected to increase in prevalence by nearly 50\% from 196 million cases worldwide in 2020 to 288 million by 2040 \citep{wong2014global}.\\
In this paper we extend contrastive pretraining with strategies leveraging time series of widely available metadata in longitudinal retinal imaging datasets (see Figure \ref{fig:study}). We conduct our analysis across seven diverse downstream tasks related to AMD in two large datasets totalling 170,427 OCT images of 7,912 patients collected in the scope of the PINNACLE study \citep{sutton2022developing}. Overall, our contributions and key findings include:
{
\begin{itemize}
    \item We adapt the set of transformations and pretraining protocols for two standard contrastive pretraining methods in datasets of retinal OCT images.
    \item We then introduce metadata-enhanced learning, leveraging widely available patient metadata to address known issues with standard contrastive methods. To this end, our approach flexibly encodes known temporal dynamics of disease progression into pretraining using widely available metadata. Specifically, we use the scan date, anonymous patient identifier and eye laterality to create more informative contrastive positive pairs that correct misleading false negative pairs that arise systematically in standard contrastive frameworks.
    \item We found that models pretrained with standard contrastive methods consistently outperformed models pretrained on ImageNet and RadImageNet, and matched a retinal foundation model trained on as many as 15x retinal OCT images. Moreover, models pretrained with metadata-enhanced learning, redefining contrastive relationships between longitudinal images acquired within 1 year intervals, consistently outperformed both standard contrastive methods and the retinal foundation model. Notably, to classify Healthy vs. Early AMD and Late vs. Early AMD metadata-enhanced extensions required 20x and 100x fewer labelled data, respectively, to recover the performance of the foundation model.
\end{itemize}
}

\section{Related Work}
\subsection{Self-supervised learning}
Self-supervised pretraining has gained momentum in recent years as a method for creating generalisable representations of unlabelled data. Pretrained models can then be finetuned on supervised tasks resulting in faster training and requiring fewer annotations to reach good performance. It works by solving so-called pretext tasks that create learning signals without human supervision. Initially pretext tasks were derived from ad-hoc heuristics such as predicting image rotation \citep{komodakis2018unsupervised} and colourisation \citep{zhang2016colorful} but these were surpassed by more general context matching tasks such as solving jigsaw puzzles \citep{noroozi2016unsupervised} and predicting one local context from another \citep{oord2018representation}.\\
Contrastive frameworks \citep{chen2020simple, he2020momentum, grill2020bootstrap, chen2021exploring} currently advance the state-of-the-art in self-supervised learning. In their seminal work, \citealt{chen2020simple} proposed SimCLR which creates a context matching task by combining a set of image transformations with a contrastive loss. They define similarity at the image level, such that two augmented views of the same image create a positive contrastive pair and views from different images make negative pairs. Models then learn to maximise similarity in representation space between positive pairs, and minimise similarity between negative pairs. They find that a diverse set of transformations, including random crops, rotations and colour shifts, helps in building more generalisable representations than could be learned from any one pretext task.\\ 
\citealt{khosla2020supervised} show that standard contrastive learning on ImageNet is degraded by misleading negative pairs consisting of images belonging to the same class, and restore performance by instead making these pairs positive. In Bootstrap Your Own Latent (BYOL) \citealt{grill2020bootstrap} remove negative pairs altogether and employ imbalanced student and teacher networks to prevent a degenerate solution. Overall, choosing an appropriate and diverse set of transformations as well as the correct handling of false negative pairs are crucial for the successful application of contrastive learning.

\begin{figure*}[t]
    \centering
    \includegraphics[width=0.99\linewidth]{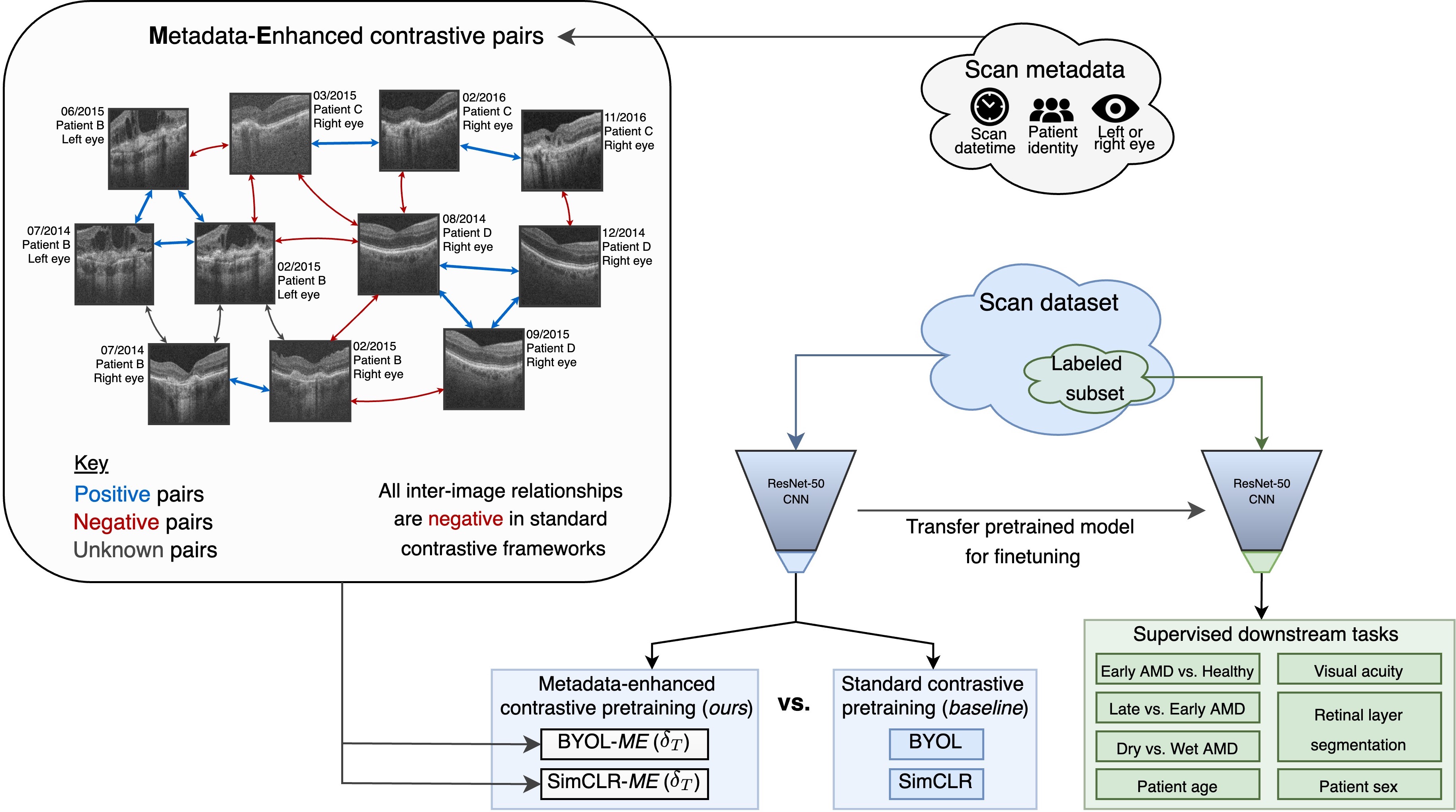}
    \caption{In this paper we pretrain models with existing contrastive frameworks (BYOL and SimCLR) and our metadata-enhanced contrastive versions on large datasets of unlabelled OCT images. Our method addresses existing weaknesses of standard frameworks by leveraging metadata widely available in the clinical workflow. To this end we employ information of patient identity, eye position (i.e. left or right) and time series information to indicate the set of true inter-image contrastive relationships. We benchmark pretraining strategies by quantifying improvements on seven downstream tasks related to the clinical assessment of AMD.}
    \label{fig:study}
\end{figure*}

\subsection{Self-supervised learning in the medical domain}
\label{related_work:self_supervised_medical}
Early applications of self-supervised learning to medical images also relied on ad-hoc pretext tasks such as image registration \citep{li2017non} or region-specific tasks such as locating anatomical landmarks in the heart \citep{bai2019self}, distance between random patches in the brain \citep{spitzer2018improving} and estimating retinal thickness in the eye \citep{holmberg2020self}. \citealt{chen2019self} use a generic context restoration task to improve plane detection in ultrasound, landmark detection in CT and tumour segmentation in MRI.\\
There are very few applications of standard contrastive learning to medical datasets \citep{ghesu2022self}. Many transformations used to generate views of natural images are not compatible with medical images. In contrast to colour images medical scans are often single-channel rendering two of the most performant augmentations in SimCLR, colour shift and greyscale, inapplicable. Additionally, medical scans often have a strong prior on their content. For example, studies typically collect scans targeting a single anatomical region or medical condition. Views generated from different images that feature similar disease-related or anatomical content will nonetheless form misleading negative pairs and degrade the learning signal.\\
To address these problems others have used domain specific information. \citealt{chen2021uscl} argue that views generated from any two frames of the same ultrasound video are highly correlated. By switching their contrastive relationship from negative to positive they show improvement in detection of COVID-19 and pneumonia. Others generate more informative positive pairs by choosing bilateral views to improve classification of conditions in dermatological photographs \citep{azizi2021big}, multi-modal views between genetic information and fundus images for classifying cardiac disease \citep{taleb2022contig}, views from the same temporal longitude for improving lung segmentation \citep{zeng2021contrastive} and views with the same disease label for classifying pleural effusion in chest X-ray \citep{vu2021medaug}. Two studies create positive pairs from spatially proximal 2D slices to improve volumetric segmentation of cardiac MRI \citep{zeng2021positional, chaitanya2020contrastive}. \citealt{ciga2022self} limit negative pairs to histopathological images with different staining and tissue types for improved detection of cancer.

\subsection{Deep learning for retinal imaging and AMD}
Most existing applications of deep learning to retinal images are fully supervised \citep{bogunovic2017machine, lee2017deep, schmidt2018prediction, de2018clinically}. There have been very few applications of self-supervised learning to retinal images. \citealt{holmberg2020self} use a multi-modal approach to improve detection of retinopathy by pretraining models to use fundus photographs to predict retinal thickness maps derived from OCT. \citealt{srinivasan2021robustness} find that contrastive pretraining on ImageNet boosted downstream analysis of diabetic retinopathy in small fundus imaging datasets.\\
Self-supervised pretraining for AMD has so far included one work using fundus photographs \citep{yellapragada2022self} and another employing a temporal ordering task to learn late AMD features from longitudes in OCT \citep{rivail2019modeling}. {More recently \citep{zhou2023foundation} released RETFound, a foundation model for retinal images trained on over 700,000 OCT images. However,} there are no studies comprehensively testing the potential role of metadata for improving self-supervised learning in the domain of retinal OCT.\\

\section{Materials and Methods}
Our proposed method is summarised in Figure \ref{fig:study}. In this section we describe the two longitudinal retinal OCT datasets in section \ref{methods:datasets}. We then revisit standard contrastive learning and introduce our proposed metadata-enhanced framework in section \ref{methods:pretrain}. Finally, we specify experiments to evaluate our pretrained models and the benefit of our metadata-enhanced modifications in section \ref{methods:experiments}.

\begin{figure*}
    \centering
    \subfloat {\includegraphics[width=0.97\linewidth]{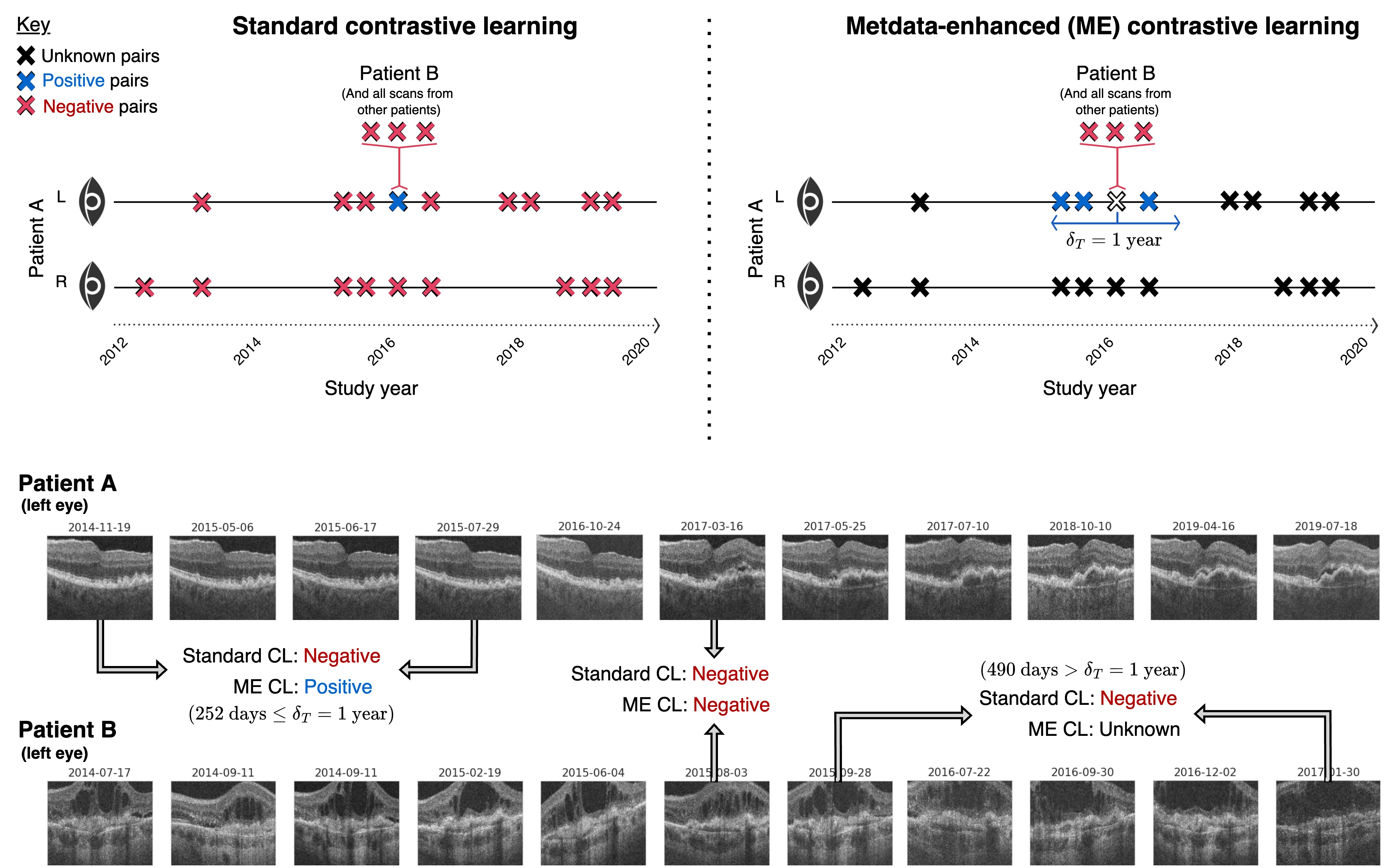}
    }
    \caption{
     Our method enhances standard contrastive pretraining with {widely available} medical metadata to correct many of the misleading negative pairs that arise {systematically} in standard frameworks. Moreover, by introducing inter-image positive pairs we combine artificial image transformations with natural ones that already exist between images acquired closely in time (controlled by a $\delta_T$ parameter). Our method also removes contrastive pairs with unknown relationships and retains negative pairs containing images originating from different patients.
    }
    \label{fig:figure2}
\end{figure*}

\begin{figure}
    \centering
    \subfloat[Distribution of longitude lengths] {\includegraphics[height=0.35\linewidth]{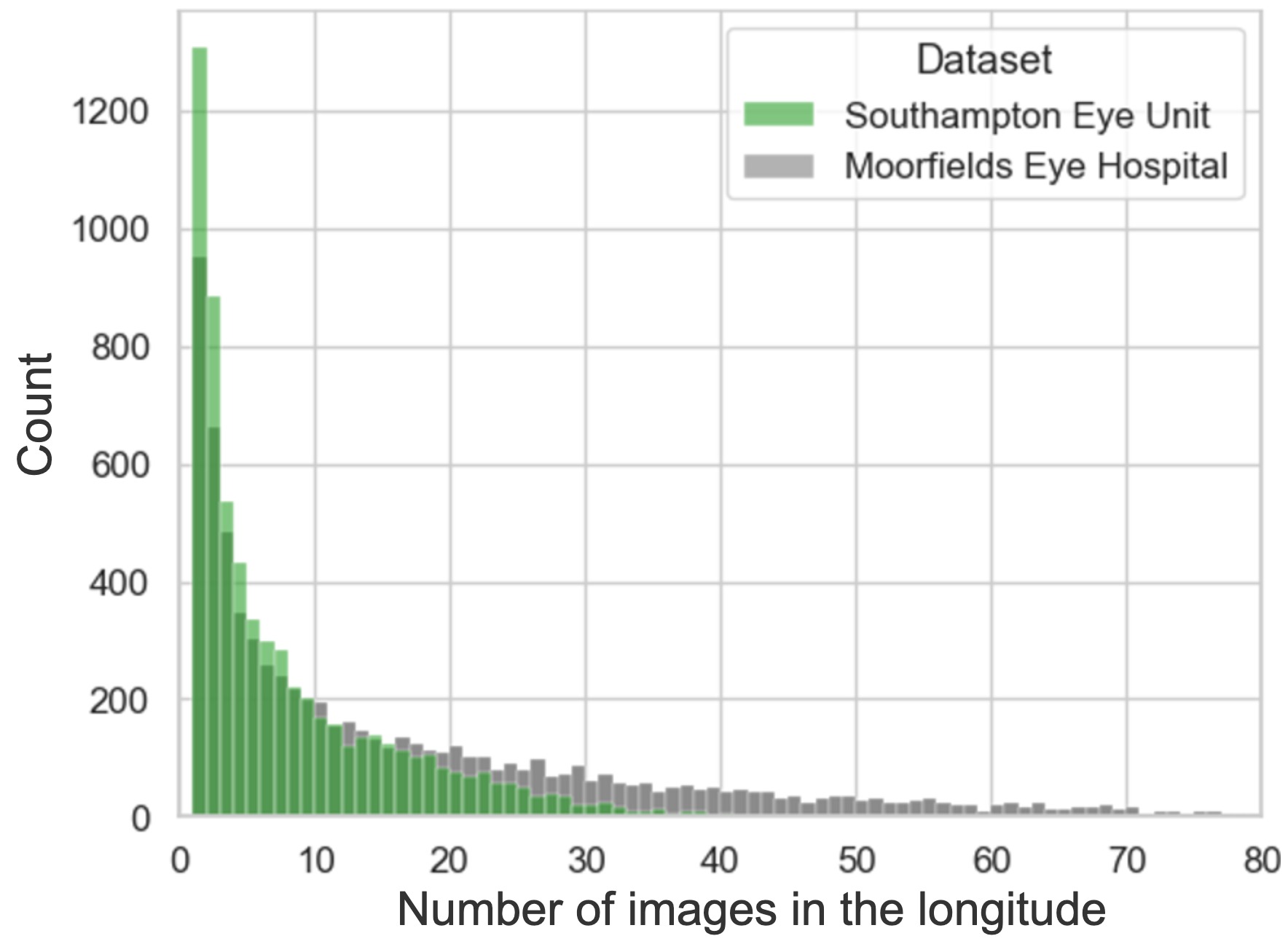}
    \label{fig:longitude_figure}
    }
    \quad
    \subfloat [Distribution of time intervals between scans of the same longitude] {\includegraphics[height=0.35\linewidth]{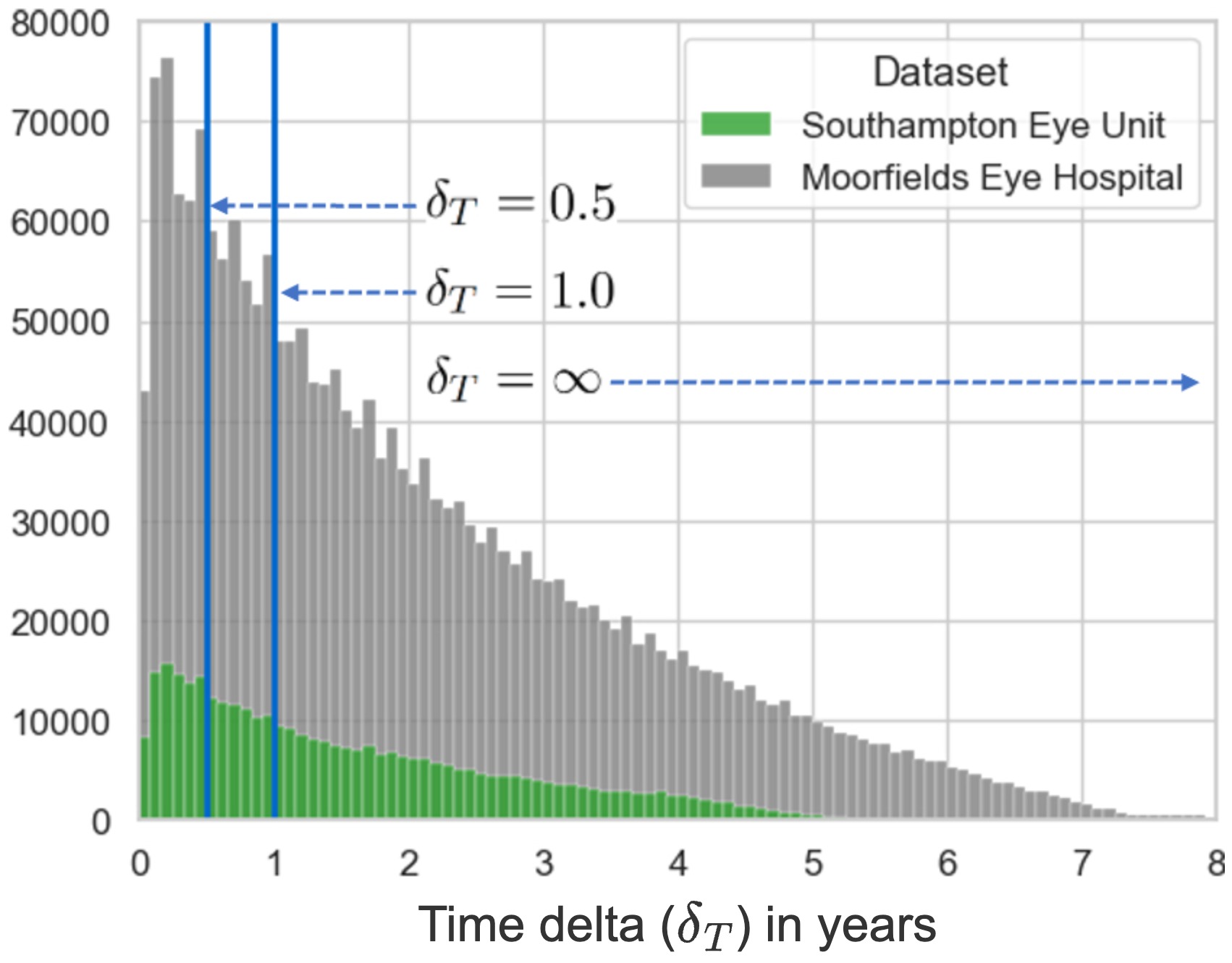}
    \label{fig:delta_t}
    }
    \quad
    \caption{The distribution of longitude lengths (left) in the Southampton and Moorfields datasets and the frequency distribution of time intervals $\delta_T$ in years between all pairs of longitudinal scans from the same eye (right), with the width of each bin covering a duration of one month.}
\end{figure}

\subsection{OCT datasets}
\label{methods:datasets}
In this paper we use two datasets collected as part of the PINNACLE project \citep{sutton2022developing}. They each contain longitudinal data consisting of retinal OCT scans of patients with AMD. For both datasets images were acquired using Topcon 3D OCT devices (Topcon Corporation, Tokyo, Japan). The first dataset, collected at the Southampton Eye Unit, consists of 48,825 OCT scans and followed 6,368 eyes from 3,498 patients. A subset of 2,037 images are healthy control scans curated from 570 patients. In Figure \ref{fig:longitude_figure} we show the number of times eyes were scanned. The average eye was scanned 7.7 times over an average duration of 1.9 years, with longitudes featuring AMD spanning 2.3 years. The second dataset, collected at Moorfields Eye Hospital, is larger containing 121,602 OCT scans of 7,336 eyes from 3,844 patients. The longitudinal dimension is also greater, with longitudes containing on average 16.6 scans spanning 3.5 years. We extracted the mediolateral 2D slice centred at the fovea and resampled to 208$\times$256 pixels with a pixel size of 7.0×23.4 $\mu m^2$, half the median resolution. {We elect to maintain these as separate datasets to test whether any benefits from our pretraining methods generalise to different hospitals and collection sites.}\\
Each image is accompanied by metadata detailing which eye (i.e. left or right eye) is depicted, the patient's anonymised identity and the exact date and time of scanning. This metadata was reserved for metadata-enhanced pretraining while other demographic information recording the patient's age and sex are instead used for downstream evaluation.\\
AMD stage labels, indicating whether a patient has early or late stage AMD, were extracted automatically from the electronic health records. For 39\% of scans it was possible to extract an AMD stage label in this manner, and the remaining 61\% were left unlabelled. Late stage AMD was further characterised into either wet (choroidal neovascularization) or dry (geographic atrophy) AMD. Clinical information also included visual acuity, a metric for assessing functional quality of vision, which measured the logarithm of the minimum angle of resolution (LogMAR) perceptible to the patient. {}{}{We rescale this measurement via a linear mapping to find the Letter Score. This is a more intuitive metric which indicates how many letters, from 0 to 95, could be read by the patient on a letter chart \citep{beck2003computerized}.} Finally, we obtained segmentations of eleven retinal layers using the Iowa Reference Algorithms (Retinal Image Analysis Lab, Iowa Institute for Biomedical Imaging, USA), a graph-cut-based segmentation algorithm \citep{li2005optimal, garvin2009automated, abramoff2010retinal}.

\subsection{Metadata-enhanced contrastive learning}
\label{methods:pretrain}
{
In this study, we first extend SimCLR \citep{chen2020simple} as it is the most widely adopted and paradigmatic contrastive method \citep{chaitanya2020contrastive,azizi2021big,ciga2022self,taleb2022contig}. For a second standard baseline we include BYOL \citep{grill2020bootstrap} for its exclusive use of positive pairs. As alluded to in section \ref{related_work:self_supervised_medical} standard contrastive learning suffers from two issues in medical applications. The first is that several transformations used by SimCLR and BYOL, such as hue, saturation and colour dropping, are not applicable in single-channel medical scans. The second is that many negative contrastive pairs will in fact feature highly similar views since all OCT images depict the same anatomy and limited set of pathologies, especially if both images originate from the same eye or patient.}\\
To address these issues we redefine contrastive inter-image relationships as depicted in Figure \ref{fig:figure2}. To this end we leverage the longitudinal medical metadata described in section \ref{methods:datasets}. Let $L$ longitudinal scans of an eye with position $l$ (i.e. left or right eye) from patient $u$ be ${}^{u}e^{l}=\{({}^{u}x^{l}_i, {}^{u}t^{l}_i)\}_{i=1}^{L}$ where $x_i$ and $t_i$ denote the scan image data and acquisition timestamp, respectively. We define two scans as being in temporal proximity if the time between their acquisition dates falls between $\delta_T^{min}$ and $\delta_T$. Finally, we define a function $S$ for the contrastive relationship between any two scans as:
\begin{align}
S_{\delta_T^{min}, \text{ } \delta_T}({}^{u}x^{l}_i, {}^{v}x^{m}_j)
&= \begin{cases}
  \boldsymbol{\pmb{+}} & (u = v) \land (l = m) \land (\delta_T^{min} \leq |t^u_i - t^v_j| \leq \delta_T)\\
\boldsymbol{\pmb{-}} & u \neq v\\
  {\boldsymbol{\pmb{?}}} & {\text{otherwise}}
\end{cases}
\label{eq:S}
\end{align}

\begin{algorithm}[ht!]
\caption{{\textbf{PyTorch Code 1}: Metadata-enhanced wrapper for standard contrastive losses. This function uses the patient identifier, eye side, and scan date of each image to construct more informative positive and negative pairs for standard contrastive methods. This customizable approach allows for flexible encoding of in-domain knowledge into pretraining. In this paper, we choose to define positive pairs as images of the same eye acquired within a short temporal interval defined by $\delta_T$. We also use metadata to exclude negative pairs of images originating from either eye of the same patient.}}
\label{alg:me}
\begin{lstlisting}[language=Python]
def standard_contrastive_loss(positive_pairs, negative_pairs):
    # Any contrastive loss such as SimCLR, BYOL, MoCo, DINO...
    ...

def metadata_enhanced_contrastive_loss(embeddings, metadata, d_T_min, d_T):
    patient_ids, eye_sides, scan_dates = metadata

    # Use metadata to identify positive and negative pairs 
    # (excluding unknown pairs by default)
    time_intervals = (scan_dates - scan_dates.T).abs()
    temporal_mask = (d_T_min < time_intervals) & (time_intervals <= d_T_max)
    same_patient_mask = (patient_ids == patient_ids.T)
    same_eye_mask = same_patient_mask & (eye_sides == eye_sides.T)

    positive_pairs = embeddings[torch.where(same_eye_mask & temporal_mask)]
    negative_pairs = embeddings[torch.where(~same_patient_mask)]

    # Calculate contrastive loss as normal on redefined pairs
    loss = standard_contrastive_loss(positive_pairs, negative_pairs)
    return loss
\end{lstlisting}
\end{algorithm}

This equation encodes three assumptions. Firstly, we assume that longitudinal images of the same eye acquired within a set time window $\delta_T$ differ minimally in their progression of AMD and healthy ageing (see Figure \ref{fig:figure2}). These cases are labelled as positive pairs for the purpose of contrastive learning. To observe evidence of this assumption refer to Figure \ref{fig:longitude_examples}. {The second assumption states that pairs of images from different individuals are likely to be dissimilar and we label these pairs as negative accordingly. Furthermore, scans outside the $\delta_T$ window may be similar in cases of slow AMD progression but semantically different in cases of fast disease progression. Consequently, we remove these as contrastive pairs altogether. Similarly, disease progression is known to be correlated in fellow eyes with a temporal delay. Therefore, we also remove pairs of images from fellow eyes due to their unknown relationship. The third assumption states that pairs of images from different individuals are more likely to be dissimilar with respect to disease state, and we label these pairs as negative accordingly.}\\
Using Equation \ref{eq:S} to redefine inter-image relationships, we introduce metadata-enhanced (ME) variants of BYOL and SimCLR as BYOL-ME($\delta_T$) and SimCLR-ME($\delta_T$), respectively. To create augmented views we retain the set of transformations used in standard contrastive learning. However, instead of creating two views from each image we now generate positive pairs from two related images by augmenting each once. When creating batches we ensure that every image is part of at least one positive pair. {While we chose SimCLR and BYOL in this paper as two of the most widely adopted and researched contrastive frameworks, our positive and negative pair reformulation is applicable to any contrastive framework and can be calibrated for any progressive disease.}

Our metadata-enhanced framework can be implemented by a simple extension to existing contrastive losses (see PyTorch Code \ref{alg:me}). By redefining positive and negative pairs using a customisable set of rules, existing pipelines can be modified at minimal cost.

\newpage
\subsection{Experimental protocol}
\label{methods:experiments}
\subsubsection{Contrastive pretraining protocol}
\label{methods:pretraining_protocol}
{}{}{Both standard and metadata-enhanced contrastive pretraining variants, for both SimCLR, BYOL, use a ResNet50 (4x) backbone with half precision, and were randomly initialised before training according to the He protocol \citep{he2016deep}}. In both datasets we allocate 4,800 samples for validation, 4,800 for testing and the remainder for pretraining. {}{}{Models then train with the AdamW optimiser using a learning rate of $5\cdot10^{-4}$ and momentum of $0.9$ for a fixed number of 120,000 optimisation steps}. We use linear warmup for the first 1,200 steps and decay the learning rate with the cosine schedule without restarts. The batch size is 384 and we use the model after the final training step for downstream evaluation. Additional details are listed in \ref{appendix:pretraining}. Due to the sensitivity of contrastive frameworks to the chosen transformations we closely follow those used by BYOL \citep{grill2020bootstrap}, transferring as many as are applicable to retinal OCT. We first, with probability 0.8, augment each scan by varying image brightness and contrast using maximum relative changes of 0.4. Then we rotate by a random angle up to 8 degrees, crop centrally to resolution 188$\times$236 to remove rotation border artefacts, flip horizontally with probability 0.5, randomly crop to a scale between 0.25 and 1.0 with aspect ratio between 3/4 and 4/3, and finish by resizing to 192$\times$192. {We chose not to use Gaussian blur as we, in accordance with \citep{azizi2021big}, found it obfuscated local textural features in the retinal images.} We also omit hue, saturation and colour dropping as these augmentations do not have any effect on greyscale images.\\
In our experiments we first pretrain multiple ResNet50 models using both standard SimCLR and BYOL in addition to their metadata-enhanced extensions defined in section \ref{methods:pretrain}. For both SimCLR-ME and BYOL-ME we test three values of $\delta T$: 0.5 years, 1.0 years and no time limit (i.e. all available longitudinal scans). These values result in including 18.8\%, 35.1\% and 100\% of the potentially positive pairs respectively, or 469,334, 874,029 and 2,493,568 in total as can be seen in Figure \ref{fig:delta_t}. We set $\delta_T^{min}$ to 0.02 years to exclude poor quality pairs that are likely due to a scan retake.

{
\subsubsection{Pretrained baselines}
\label{sec:pretrained_baselines}
To fairly assess the benefit of metadata-enhanced learning we compare against three pretrained baselines. The first baseline is a model pretrained by \citep{chen2020simple} in SimCLR on natural images in ImageNet. Our second baseline uses a more medically aligned model pretrained to classify 165 different conditions on RadImageNet \citep{mei2022radimagenet}. We also compare our methods against RETFound, a foundation model for retinal images trained on over 700,000 OCT images \citep{zhou2023foundation}. RETFound was trained as a Masked Autoencoder \citep{he2022masked} and uses a ViT-Large backbone. We use the version specifically trained on OCT data, which was found to be highly performant in retinal disease prediction, including AMD, in external datasets. These baselines, pretrained on natural images, medical images and external retinal OCT images, offer increasing levels of in-domain pretraining against which to compare in-dataset contrastive learning, and our metadata-enhanced extensions.
}

\subsubsection{Finetuning protocol}
\label{methods:finetune}
{To compare different pretrained models we adopt the standard linear evaluation finetuning protocol used by \citep{chen2020simple}. We first freeze the pretrained encoder} and replace any dense projection layers used in pretraining with a single linear layer that projects from the latent space to the dimension of the supervised label. More specifically, for the segmentation task this involves a 11x192 dimension matrix and for all other problems the output is size 1. For classification problems we then apply a sigmoid activation function and for segmentation and regression problems we use $\tanh$ activation and scale labels to the range [-1,1]. {We use a batch size of 1024 and ensure we use the same train, validation and test splits as for pretraining. We then train all models for 3000 epochs using the early stopping if downstream validation performance does not improve for 600 steps. To fit the model we use an AdamW optimiser with momentum of $0.9$, weight decay of $1 \cdot 10^{-2}$ and a learning rate of $5\cdot10^{-4}$ using the cosine annealing schedule after 50 linear warmup epochs. These hyperparameters are the same as were used to evaluate RETFound in \citep{he2022masked}. Moreover, for all models pretrained on external dataset we also resize the input image to the size they were trained on, which was 224x224.} In training we employ a weaker form of the augmentations described in the pretraining protocol. Finally, we report performance on the test set of the model checkpoint with the best validation downstream performance.

\subsubsection{Downstream tasks}
\label{methods:downstream}
To evaluate pretrained models, including pretrained baselines and models pretrained with standard and metadata-enhanced contrastive learning, we test them on the following seven downstream tasks:
\begin{enumerate}
    \item \textbf{AMD stage and type classification tasks} to test the ability of pretrained models to learn disease progression and disease type. The following tasks are tested: 
    \begin{enumerate}
 \item Early/Intermediate AMD vs. Healthy eye classification. 
 \item Late vs. Early/Intermediate AMD classification.
 \item Late stage Dry vs. Wet AMD classification. 
    \end{enumerate}
    \item \textbf{Prediction of functional vision assessment score}, which involves learning any clinical features visible in the retina that may degrade the patient's quality of vision.
    \item \textbf{Prediction of patient demographic information}, including patient sex classification and age regression to incorporate learning healthy ageing and demographic characteristics.
    \item {\textbf{Dense retinal surface segmentation}, where models simultaneously regress the column-wise location of eleven retinal cell layers along each OCT A-scan as in \citep{he2021structured}. {Unlike the aforementioned image-level tasks, this task requires learning pixel-level structural variations} which are known to be affected by pathology, healthy ageing and differences in patient sex.}
\end{enumerate}

For all tasks we tested scenarios where fewer labelled samples are available for training. To do this we sample exponentially increasing amounts of labelled data for training, starting from just 20 training samples and increasing to 10,000 (for subset selection protocol refer to \ref{appendix:subsets}). Performance on classification tasks is measured in area under the receiver operator characteristic curve (AUC) and regression and segmentation tasks in mean average error (MAE). {We repeat all downstream experiments using \K different seeds and report performance in a low-data regime finetuned on 100 labelled samples, or 1\% of the data, against a full high-data regime using a maximum of 10,000 labelled samples.} Our entire workflow and results are repeated across both retinal OCT datasets. In total our results summarise 4,620 finetuning runs evaluating 11 different pretraining strategies.

\newpage
\section{Results}
\begin{figure}[!htb]
    \centering
  \includegraphics[width=0.99\linewidth]{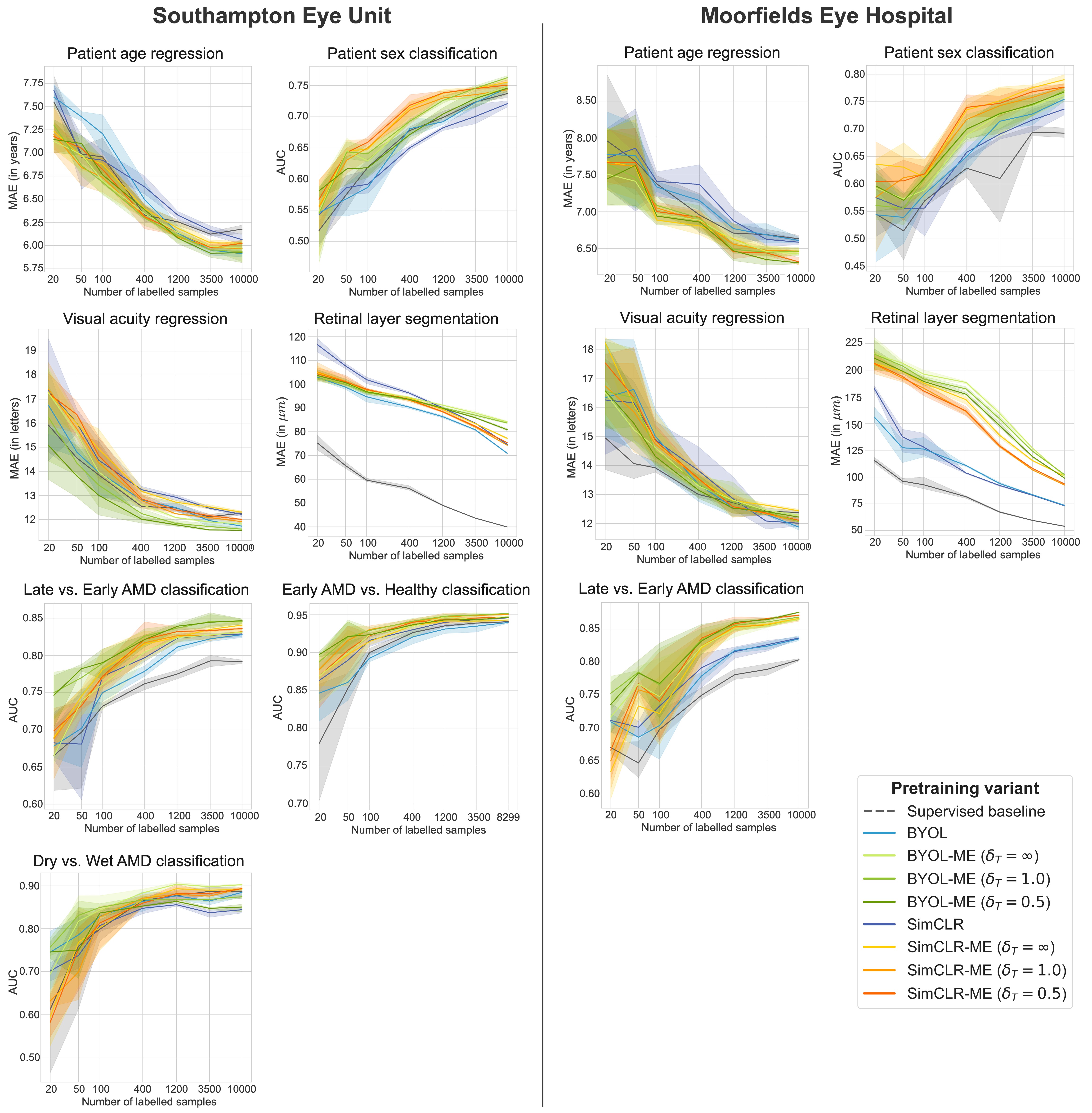}

\caption{Results of {linear evaluation} on downstream tasks using seven logarithmically spaced amounts of labelled finetuning samples (with 95\% confidence intervals). {For clarity we omit the ImageNet and RadImageNet models, which were consistently outperformed by the RETFound baseline} (model colour key in bottom right). {In all tasks except segmentation, and visual acuity on Moorfields data, metadata-enhanced contrastive pretraining extending BYOL and using $\delta_T\leq1.0$ outperforms standard contrastive learning and RETFound, especially in scenarios with fewer labelled data.}}
\label{fig:linear_eval}
\end{figure}

{
\subsection{Self-supervised pretraining on retinal OCT data is crucial for downstream performance}
Models pretrained on retinal OCT images markedly outperformed models pretrained on natural images such as ImageNet, as shown in Table \ref{tab:100_1000} and Figure \ref{fig:linear_eval}. Features extracted from RadImageNet, containing 1.35 million radiological images, were more transferable than those from ImageNet, but were still outperformed by all retinal OCT pretrained models. Full end-to-end finetuning (shown in Table \ref{tab:full_finetune}) enables models with out-of-domain pretraining to reduce the disparity in performance with models pretrained on OCT images. Furthermore, standard contrastive methods pretrained on our datasets match the performance of RETFound despite using 13x fewer parameters and training on as much as 15x fewer unlabelled images. For example, in age regression on the Southampton dataset, standard BYOL underperforms compared to RETFound in the low-data regime of  (7.21 vs. 6.96 MAE) but performs better in the high-data regime (5.91 vs. 6.18 MAE).\\
In linear evaluation, RETFound, which was trained as a Masked Autoencoder to reconstruct images on a patch-level, outperformed all other methods on retinal layer segmentation. This reflects other works suggesting that image-level representation learning tasks, such as contrastive learning, do not generalise well to dense pixel-level tasks like segmentation \citep{wang2021dense,li2022global,hu2021region,chaitanya2020contrastive}. 
}

\begin{table*}[ht!]
\caption{{We compare linear evaluation performance of different pretraining strategies across seven retinal downstream tasks on the Southampton and Moorfields datasets. We find metadata-enhanced pretraining strategies outperformed all other methods in all but one image-level task in both the low-data and high-data regime on both datasets. Best performance is highlighted in bold, and second best is underlined.}}
\label{tab:100_1000}
\resizebox{\textwidth}{!}{
\begin{tabular}{lr|cccccccccccccc}
\multicolumn{14}{c}{\textbf{\fontsize{14pt}{16pt}\selectfont\textbf{\SH}}} \vspace{0.03in} \\ \hline 
\multicolumn{2}{c|}{\begin{tabular}[c]{@{}c@{}}Pretraining variant\end{tabular}}  &\multicolumn{2}{c}{\begin{tabular}[c]{@{}c@{}}Early vs.\\Healthy (AUC) $\uparrow$\end{tabular}} & \multicolumn{2}{c}{\begin{tabular}[c]{@{}c@{}}Late vs. Early\\(AUC) $\uparrow$\end{tabular}} & \multicolumn{2}{c}{\begin{tabular}[c]{@{}c@{}}Visual acuity\\(MAE Letters)$\downarrow$\end{tabular}} &  \multicolumn{2}{c}{\begin{tabular}[c]{@{}c@{}}Dry vs. Wet\\(AUC) $\uparrow$\end{tabular}} & \multicolumn{2}{c}{\begin{tabular}[c]{@{}c@{}}Patient Age\\(MAE years)$\downarrow$\end{tabular}}  & \multicolumn{2}{c}{\begin{tabular}[c]{@{}c@{}}Patient Sex\\ (AUC) $\uparrow$\end{tabular}} & \multicolumn{2}{c}{\begin{tabular}[c]{@{}c@{}}Segmentation\\(MAE $\mu$m) $\downarrow$\end{tabular}}  \\
\multicolumn{2}{c|}{\# finetuning labels} & 100& 8299& 100& 10000 & 100& 10000& 100 & 10000  & 100& 10000 & 100& 10000  & 100 & 10000  \\ \hline
\multicolumn{2}{r|}{ImageNet SimCLR}& 0.551  & 0.587  & 0.568 & 0.537  & 18.4  & 18.5& 0.570 & 0.620 & 7.50 & 7.51& 0.516  & 0.518  & 98.3& 86.4\\
\multicolumn{2}{r|}{RadImageNet pretrained}& 0.775  & 0.712  & 0.676 & 0.745  & 20.7  & 15.4& 0.480 & 0.536& 7.75& 6.97& 0.529  & 0.627  & 177& \underline{62.0}  \\
\multicolumn{2}{r|}{RETFound  (foundation model)} & 0.900 & 0.941  & 0.732 & 0.792  & 13.9  & 12.3& 0.798& 0.885& 6.96& 6.18& 0.619  & 0.737  & \textbf{59.8} & \textbf{40.1} \\ \hline
SimCLR & Standard  & 0.916  & 0.946  & 0.772 & 0.829  & 14.5  & 12.2& 0.807& 0.843& 6.92& 6.06& 0.592  & 0.721  & 102& 74.6\\
BYOL& Standard  & 0.892  & 0.940& 0.750  & 0.828  & 14.0 & 11.7& 0.828& 0.883& 7.21& \textbf{5.91} & 0.587  & 0.746  & \underline{94.4}& 70.8\\ \hline
\multirow{3}{*}{SimCLR} & \ME{\infty}& 0.915  & 0.943  & 0.781 & 0.841  & 14.9  & 12.3& 0.802& 0.891& 6.88& 6.03& 0.648  & \underline{0.755}& 97.3& 76.8\\
& \ME{1.0} & \textbf{0.930} & \underline{0.950} & 0.770  & 0.832  & 14.4  & 11.9& 0.827& \underline{0.893}& \underline{6.74}& 5.97& \underline{0.65} & 0.743  & 98.3& 74.2\\
& \ME{0.5} & 0.922  & 0.946  & 0.772 & 0.836  & 14.7  & 12.0 & 0.813& 0.892& 6.82& 6.02& \textbf{0.659}& 0.751  & 98.0 & 75.5\\ \hline
\multirow{3}{*}{BYOL}& \ME{\infty}& 0.923  & 0.947  & 0.761 & 0.831  & 14.0& 11.8& \underline{0.842}& \textbf{0.901} & \underline{6.70} & 5.97& 0.629  & 0.748  & 96.7& 84.2\\
& \ME{1.0} & \underline{0.929}& \textbf{0.951}& \underline{0.789}  & \underline{0.845}& \underline{13.4}& \underline{11.6}& \textbf{0.849} & 0.873& \textbf{6.68} & 5.93& 0.641  & \textbf{0.763}& 96.1& 83.7\\
& \ME{0.5} & 0.923  & 0.946  & \textbf{0.790}& \textbf{0.847}& \textbf{13.0}  & \textbf{11.5} & 0.836& 0.849& 6.76& \underline{5.92}& 0.618  & 0.746  & 96.3& 81.3  

  \\ \\

\multicolumn{14}{c}{\textbf{\fontsize{14pt}{16pt}\selectfont\textbf{\MF}}} \vspace{0.03in} \\ \hline 
\multicolumn{2}{r|}{ImageNet SimCLR}& --- & --- & 0.520& 0.460  & 17.7& 17.8  & ---& ---  & 8.24& 8.17& 0.479& 0.468 & 165 & 108  \\
\multicolumn{2}{r|}{RadImageNet pretrained}& --- & --- & 0.549  & 0.524 & 16.7& 15.5  & ---& ---  & 7.57& 7.27& 0.497& 0.559 & 153 & 88.1 \\
\multicolumn{2}{r|}{RETFound  (foundation model)} & --- & --- & 0.698  & 0.804 & \textbf{13.9} & 12.4  & ---& ---  & 7.38& 6.63& 0.569& 0.693 & \textbf{89.3} & \textbf{54.0} \\ \hline
SimCLR & Standard  & --- & --- & 0.733  & 0.836 & 14.9& 12.0& ---& ---  & 7.41& 6.59& 0.556& 0.736 & 122 & \underline{73.0} \\
BYOL& Standard  & --- & --- & 0.704  & 0.835 & 15.0 & \textbf{11.9}& ---& ---  & 7.34& 6.61& 0.582& 0.754 & \underline{121}  & 73.3 \\ \hline
\multirow{3}{*}{SimCLR} & \ME{\infty}& --- & --- & 0.722  & 0.866 & 14.8& 12.4  & ---& ---  & \textbf{6.89} & 6.46& 0.612& \textbf{0.79}& 185 & 101  \\
& \ME{1.0} & --- & --- & 0.748  & 0.864 & 14.7& 12.1  & ---& ---  & 7.05& 6.47& \textbf{0.618} & \underline{0.776}  & 179 & 92.6 \\
& \ME{0.5} & --- & --- & 0.740& \underline{0.871}  & 14.8& 12.1  & ---& ---  & 7.00  & \underline{6.32}& \underline{0.618} & 0.776 & 176 & 93.5 \\ \hline
\multirow{3}{*}{BYOL}& \ME{\infty}& --- & --- & 0.748  & 0.865 & 14.6& 12.3  & ---& ---  & 6.97& 6.43& 0.589& 0.762 & 194 & 101  \\
& \ME{1.0} & --- & --- & \underline{0.765}& 0.868 & 14.5& \underline{12.1}& ---& ---  & 7.09& 6.46& 0.610 & 0.770  & 190 & 102  \\
& \ME{0.5} & --- & --- & \textbf{0.768}& \textbf{0.876}& \underline{14.3}& 12.2  & ---& ---  & \underline{6.94}& \textbf{6.31} & 0.612& 0.767 & 188 & 98.8
\end{tabular}
}
\end{table*}


\newcommand{\mpw}{0.24}
\newcommand{\gw}{0.9}

\newcommand{\fontsmall}{\fontsize{9pt}{6pt}\selectfont}

\newpage
{
\subsection{Metadata-enhanced pretraining outperforms all other forms of pretraining on image-level retinal tasks}
{We find that metadata-enhanced BYOL, which does not use any negative pairs, matched or outperformed metadata-enhanced SimCLR in all scenarios (except Sex classification in both datasets).} This leads us to conclude that misleading negative pairs still degrade performance, despite the removal of many using metadata. Furthermore, for BYOL we find that $\delta_T\leq1.0$ typically performs better than $\delta_T=\infty$. For example, BYOL-ME($\delta_T\leq1.0$) surpasses BYOL-ME($\delta_T=\infty$) in the both low and high-data regimes for Late vs. Early classification in both datasets. These results support our hypothesis that defining positive pairs over short time intervals is beneficial for image-level retinal analysis.\\
Overall, BYOL-ME $\delta_T\leq1.0$ outperforms all pretrained baselines and standard contrastive learning approaches on every image-level task in the Southampton dataset, and all but one in the Moorfields dataset. For example, to classify Late vs. Early AMD on the Moorfields dataset we are able to surpass the performance of RETFound finetuned on all 10,000 labelled samples (0.80 AUC) using only 400 samples, or 2.5\% (0.83 AUC). Similarly, we need only 100, or 1\%, to recover the same performance on the Southampton dataset (0.79 AUC). Given all 10,000 labelled samples, our approach surpasses all baselines by achieving 0.85 AUC and 0.88 AUC in the Southampton and Moorfields datasets, respectively.}

\section{Discussion}
In this paper we demonstrated the benefits of incorporating widely available metadata into self-supervised contrastive learning. {Our approach uses metadata to address known issues of contrastive methods. To this end, we used metadata indicate the true set of inter-image contrastive relationships in two longitudinal datasets of unlabelled OCT images. Our metadata-enhanced approach surpassed both standard contrastive learning and RETFound, the foundation model for retinal OCT, in both low-data and high-data regimes. Our most performant variant, BYOL-ME($\delta_T\leq1.0$), surpassed all pretrained baselines in all image-level tasks on the Southampton dataset, and all but one on the Moorfields dataset. Our strategy also enabled models using 100x fewer labelled samples to achieve the same performance as the most performant pretrained baselines in classifying Late vs. Early AMD. These results demonstrate that conditioning pretraining with widely available, yet typically unused, metadata can address known weaknesses of established contrastive frameworks in medical domains.}\\
Our study represents the first to fully establish the wide ranging potential of metadata in self-supervised pretraining in retinal OCT. Previous works in medical domains \citep{chen2021uscl, azizi2021big, taleb2022contig, zeng2021contrastive, vu2021medaug} have either focussed on evaluating multiple downstream tasks or on varying labelled subset sizes. Our work uses seven tasks, two datasets and seven labelled subset sizes to provide a comprehensive picture of {in-domain self-supervised pretraining}. Furthermore while related works focused on improving a single contrastive loss, we used both SimCLR and BYOL to ensure that pretraining with metadata represents a general improvement to constrastive learning as a whole, rather than addressing the flaws of a single framework. Overall, our experiments more closely resemble those used in studies on natural images where more exhaustive evaluations are standard practise.
\\
While our metadata-enhanced framework indicates the true contrastive relationship between two images, there exist edge cases where it will make incorrect assignments. Without access to the downstream task labels, misleading contrastive pairs containing images from different patients that, by chance, exhibit the same severity and type of AMD will not become positive pairs using our approach. Moreover, using a fixed temporal cutoff $\delta_T$ to determine positive pairs cannot encode variable rates of disease progression between eyes and individuals. A more informative set of metadata might allow for accurate negative and positive pairs between fellow eyes. This may lead to further boost to performance in smaller datasets, but in ours these pairs constitute only 0.022\% of the total. Moreover, for segmentation tasks that require encoding of local image features contrastive frameworks using high-level context matching tasks had limited value for the pixel-level segmentation task.
\\
To address some of these issues future work could, for each contrastive pair of images, compare the auto-generated metadata relationship reported by Equation \ref{eq:S} to a fully unsupervised similarity score computed using standard contrastive pretraining. Then, by highlighting cases where both BYOL and SimCLR disagree with Equation \ref{eq:S} we could begin to identify any systematic errors in our formulation. {}{}{Furthermore, future work can begin to investigate the potentially utility to prognostic predictions of disease progression through embedding temporal dynamics into pretraining \cite{shen2024spatiotemporal}}.

{
\section{Conclusion}
In this work we comprehensively demonstrated the benefits of incorporating medical metadata into self-supervised contrastive learning. Our metadata-enhanced extension of existing contrastive frameworks addresses known issues with contrastive methods while allowing researchers to flexibly encode the temporal dynamics of disease progression into pretraining. Using this approach we outperformed both standard contrastive pretraining and a foundation model for retinal images across two datasets on six diverse image-level tasks in both a low-data and high-data regime. We found benefits in downstream tasks relevant to the screening and management of AMD in retinal OCT, ranging from classification of AMD stage and type to estimation of functional endpoints. These promising findings motivate a shift in focus towards conditioning contrastive learning on medical datasets with other modalities, which can enable accurate retinal image analysis using few, gold-standard annotations.
}

\section{Acknowledgements}
We thank J. Sutton, A.J. Cree and M. Ruddock for their administrative support of this work. Funding: The PINNACLE Consortium is funded by a Wellcome Trust Collaborative Award, “Deciphering AMD by deep phenotyping and machine learning (PINNACLE)”, ref. 210572/Z/18/Z.

\appendix

\newpage
\bibliographystyle{model2-names.bst}\biboptions{authoryear}
\bibliography{custom}





\section{Supplementary material}

{
\subsection{Unfrozen finetuning results}
To test finetuning performance with unfrozen pretrained weights we use the same evaluation protocol described in Section \ref{methods:finetune} but lower the batch size to 384 and weight decay to $1.5\cdot10^{-6}$. We find unfreezing can allow inferior pretraining strategies to catch up to stronger models during finetuning, and also improve performance on retinal layer segmentation  (see Table \ref{tab:full_finetune} and Figure \ref{fig:full_finetune}). However, during full finetuning, while all other models effectively converged, RETFound faced specific challenges in convergence likely due to its use of ViT-Large. Ultimately, for RETFound we found better performance using the linear evaluation protocol. Overall, finetuning most in-domain pretrained models with unfrozen weights performed comparably or slightly worse on image-level tasks than with the linear evaluation protocol. 

\begin{table*}[ht!]
\caption{{Downstream performance of unfrozen pretrained models. In most cases models performed comparably to the frozen linear evaluation. Best performance is highlighted in bold, and second best is underlined.}}
\label{tab:full_finetune}
\resizebox{\textwidth}{!}{
\begin{tabular}{lr|cccccccccccccc}
\multicolumn{16}{c}{\textbf{\fontsize{14pt}{16pt}\selectfont\textbf{\SH}}} \vspace{0.03in} \\ \hline
\multicolumn{2}{c|}{\begin{tabular}[c]{@{}c@{}}Pretraining\\ variant\end{tabular}} & \multicolumn{2}{c}{\begin{tabular}[c]{@{}c@{}}Early vs. Healthy\\(AUC) $\uparrow$\end{tabular}} & \multicolumn{2}{c}{\begin{tabular}[c]{@{}c@{}}Late vs. Healthy\\(AUC) $\uparrow$\end{tabular}} & \multicolumn{2}{c}{\begin{tabular}[c]{@{}c@{}}Visual acuity\\(MAE LogMAR)$\downarrow$\end{tabular}} & \multicolumn{2}{c}{\begin{tabular}[c]{@{}c@{}}Dry vs. Wet\\(AUC) $\uparrow$\end{tabular}} & \multicolumn{2}{c}{\begin{tabular}[c]{@{}c@{}}Patient Age\\(MAE years)$\downarrow$\end{tabular}} & \multicolumn{2}{c}{\begin{tabular}[c]{@{}c@{}}Patient Sex\\ (AUC) $\uparrow$\end{tabular}} & \multicolumn{2}{c}{\begin{tabular}[c]{@{}c@{}}Segmentation\\(MAE $\mu$m) $\downarrow$\end{tabular}}\\
\multicolumn{2}{r|}{\#finetuning labels} & \multicolumn{1}{c}{100}& \multicolumn{1}{c}{8299} & \multicolumn{1}{c}{100} & \multicolumn{1}{c}{10000}& \multicolumn{1}{c}{100}& \multicolumn{1}{c}{10000} & \multicolumn{1}{c}{100}& \multicolumn{1}{c}{10000}& \multicolumn{1}{c}{100}& \multicolumn{1}{c}{10000}& \multicolumn{1}{c}{100} & \multicolumn{1}{c}{10000} & \multicolumn{1}{c}{100} & \multicolumn{1}{c}{10000} \\ \hline
\multicolumn{2}{r|}{No pretraining}   & 0.835 & 0.917 & 0.682 & 0.808& 16.3& 12.4 & 0.694  & 0.825  & 8.17  & 9.55  & 0.594  & 0.701  & 31.4& 17.2  \\
\multicolumn{2}{r|}{ImageNet SimCLR}  & 0.841 & 0.902 & 0.694 & 0.796& 16.2& 12.4 & 0.708  & 0.833  & 7.45  & 6.15  & 0.612  & 0.708  & 30.7& 17.9  \\
\multicolumn{2}{r|}{RadImageNet pretrained}   & 0.879 & \underline{0.944}   & 0.662 & 0.804& 18.1& 12.1 & 0.709  & \underline{0.871}& 7.40   & 6.10   & 0.562  & 0.723  & 69.3& 16.4  \\
\multicolumn{2}{r|}{RETFound  (foundation model)} & 0.919 & 0.935 & 0.748 & 0.771& 13.4& 12.6 & 0.749  & 0.812  & \textbf{7.18}  & 6.78  & 0.490  & 0.659  & \textbf{22.9} & 28.9\\ \hline
SimCLR & Standard & 0.903 & 0.941 & 0.771 & \textbf{0.849}   & 13.6& 11.7 & 0.778  & 0.860   & 7.35  & 6.07  & 0.608  & 0.742  & 32.3& 16.0\\
BYOL   & Standard & 0.903 & 0.936 & 0.754 & 0.835& 13.4& \textbf{11.5}& 0.763  & 0.869  & 7.64  & 6.08  & 0.626  & 0.730   & \underline{29.2}  & 16.3  \\ \hline
\multirow{3}{*}{SimCLR}& \ME{\infty}  & 0.905 & 0.942 & 0.779 & 0.836& 14.0  & 11.6 & 0.791  & 0.859  & 7.39  & \textbf{5.98} & 0.634  & \textbf{0.763} & 36.8& \underline{15.8}\\
   & \ME{1.0}& \textbf{0.929}& 0.938 & 0.765 & 0.838& 13.8& \underline{11.5}   & 0.725  & 0.849  & 7.31  & \underline{5.99}& 0.633  & 0.746  & 36.1& 16.1  \\
   & \ME{0.5}& 0.914 & 0.941 & 0.772 & 0.827   & 14.0  & 11.6 & 0.784  & \textbf{0.874} & 7.45  & 6.08  & \underline{0.638}& \underline{0.761}& 36.0  & 16.1  \\ \hline
\multirow{3}{*}{BYOL}  & \ME{\infty}  & 0.915 & 0.937 & \textbf{0.783}& \underline{0.845}  & 13.2& 11.8 & \textbf{0.821} & 0.864  & 7.25  & 6.16  & 0.637  & 0.748  & 34.5& 16.0\\
   & \ME{1.0}& \underline{0.924}   & 0.937 & \underline{0.779}   & 0.835& \underline{13.1}  & 11.6 & \underline{0.817}& 0.841  & 7.24& 6.16  & \textbf{0.639} & 0.733  & 34.1& 15.9  \\
   & \ME{0.5}& 0.920  & \textbf{0.948}& 0.765 & 0.832& \textbf{13.0} & 11.7 & 0.802  & 0.853  & \underline{7.19} & 6.11  & 0.611  & 0.731  & 35.6& \textbf{15.8} \\ \\

\multicolumn{16}{c}{\textbf{\fontsize{14pt}{16pt}\selectfont\textbf{\MF}}} \vspace{0.03in} \\ \hline

\multicolumn{2}{r|}{No pretraining}   & ---   & ---   & 0.639 & 0.809& 16.2 & 12.8& ---   & --- & 8.09   & 6.90  & 0.511  & 0.684  & 41.5  & 23.5  \\
\multicolumn{2}{r|}{ImageNet SimCLR}  & ---   & ---   & 0.592 & 0.815& 15.9 & 12.7& ---   & --- & 8.16   & 6.90  & 0.494  & 0.682  & 41.2   & 25.2  \\
\multicolumn{2}{r|}{RadImageNet pretrained}   & ---   & ---   & 0.607 & 0.819& 15.3 & 12.3& ---   & --- & 7.46   & 6.72 & 0.514  & 0.672  & 63.1& 22.1  \\
\multicolumn{2}{r|}{RETFound  (foundation model)} & ---   & ---   & 0.632 & 0.706& 14.3 & 13.0  & ---   & --- & 7.62   & 7.13 & 0.531  & 0.635  & \textbf{36.2} & 39.2  \\ \hline
SimCLR & Standard & ---   & ---   & 0.725 & 0.835& 13.7 & 11.7& ---   & --- & 7.24   & 6.68 & 0.588  & 0.744  & 42.3& 21.2  \\
BYOL   & Standard & ---   & ---   & 0.710  & 0.828& \textbf{13.5}& 11.8& ---   & --- & 7.24   & \textbf{6.40}  & 0.573  & 0.749  & \underline{40.5}  & 21.5  \\ \hline
\multirow{3}{*}{SimCLR}& \ME{\infty}  & ---   & ---   & 0.711 & 0.849& 14.0   & \underline{11.7}  & ---   & --- & 7.08   & 6.57 & \textbf{0.627} & \underline{0.771}& 53.4& 21.1  \\
   & \ME{1.0}& ---   & ---   & 0.722 & 0.854& 14.0   & 11.9& ---   & --- & \textbf{7.05}  & 6.55 & \underline{0.616}& 0.770   & 50.6& \underline{21.0}  \\
   & \ME{0.5}& ---   & ---   & \underline{0.739}   & \textbf{0.858}   & \underline{13.7}   & \textbf{11.6}   & ---   & --- & 7.18   & 6.55 & 0.605  & 0.755  & 48.7& 21.1  \\ \hline
\multirow{3}{*}{BYOL}  & \ME{\infty}  & ---   & ---   & 0.719 & 0.849& 13.9 & 11.9& ---   & --- & \underline{7.06} & \underline{6.50} & 0.582  & 0.745  & 47.7& 21.2  \\
   & \ME{1.0}& ---   & ---   & 0.734 & 0.851& 13.9 & 11.9& ---   & --- & 7.16   & 6.52   & 0.592  & \textbf{0.771} & 47.4& \textbf{20.9} \\
   & \ME{0.5}& ---   & ---   & \textbf{0.741}& \underline{0.856}  & 13.8 & 11.9& ---   & --- & 7.14 & 6.55 & 0.594  & 0.756  & 46.2& 21.1 
\end{tabular}
}
\end{table*}




\begin{figure}[!htb]
    \centering
\includegraphics[width=0.99\linewidth]{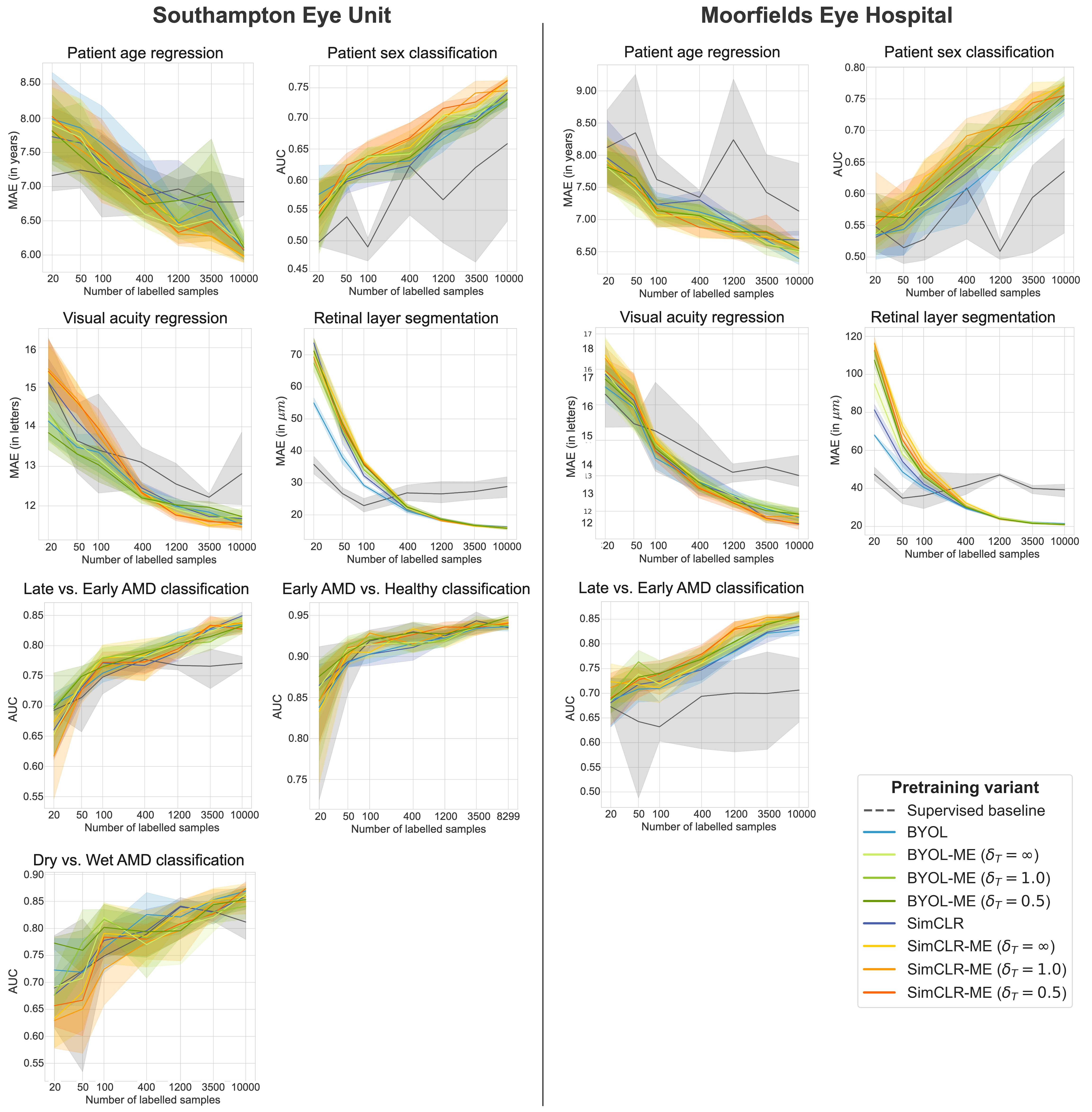}
\caption{Performance {of finetuning models with fully unfrozen weights} on downstream tasks on both datasets. Depicted is the performance (with 95\% CIs) against varying sizes of the labelled subsets used for finetuning.}
\label{fig:full_finetune}
\end{figure}

}

\subsection{Contrastive pretraining details}
\label{appendix:pretraining}
In BYOL we choose the update coefficient $\tau=0.9995$ in accordance with the authors recommendations for smaller batch sizes. We also found that for metadata-enhanced SimCLR setting $Q=20$ using the configuration described in \citep{chuang2020debiased} increased stability during pretraining.

\subsection{Finetuning details}
Pretraining speeds up the rate of convergence during finetuning. Accordingly, we found that on some tasks most pretrained models achieve optimal performance very early during finetuning. To address this validation epochs are initially run frequently at every four training steps. We then scheduled this interval to increase to 2 epochs at 100 steps, 3 epochs at 500 steps, and 5 epochs at 1000 steps to improve efficiency by skipping redundant checks later in training.

\subsection{Sampling and training on labelled subsets}
\label{appendix:subsets}
Labelled subsets for finetuning have exponentially increasing sizes. To choose these sizes we round down the number of labelled samples to the nearest 10 labels if the amount is less than 100, and we otherwise round down to the nearest 100. The largest amount is not changed so as to estimate performance with 100\% of the available labels.\\
For classification tasks we stratify our labelled subsets by uniformly sampling a proportionate amount of scans from each label class. This prevents the labelled subset from featuring only one class or exhibiting a class balance unrepresentative of the original population. Finally to enable fair comparison between different models we ensure that each is finetuned on the same set of randomly drawn labelled subsets.

\subsection{{Robustness tests}}
\begin{figure*}[t]
    \centering
    \includegraphics[width=0.9\linewidth]{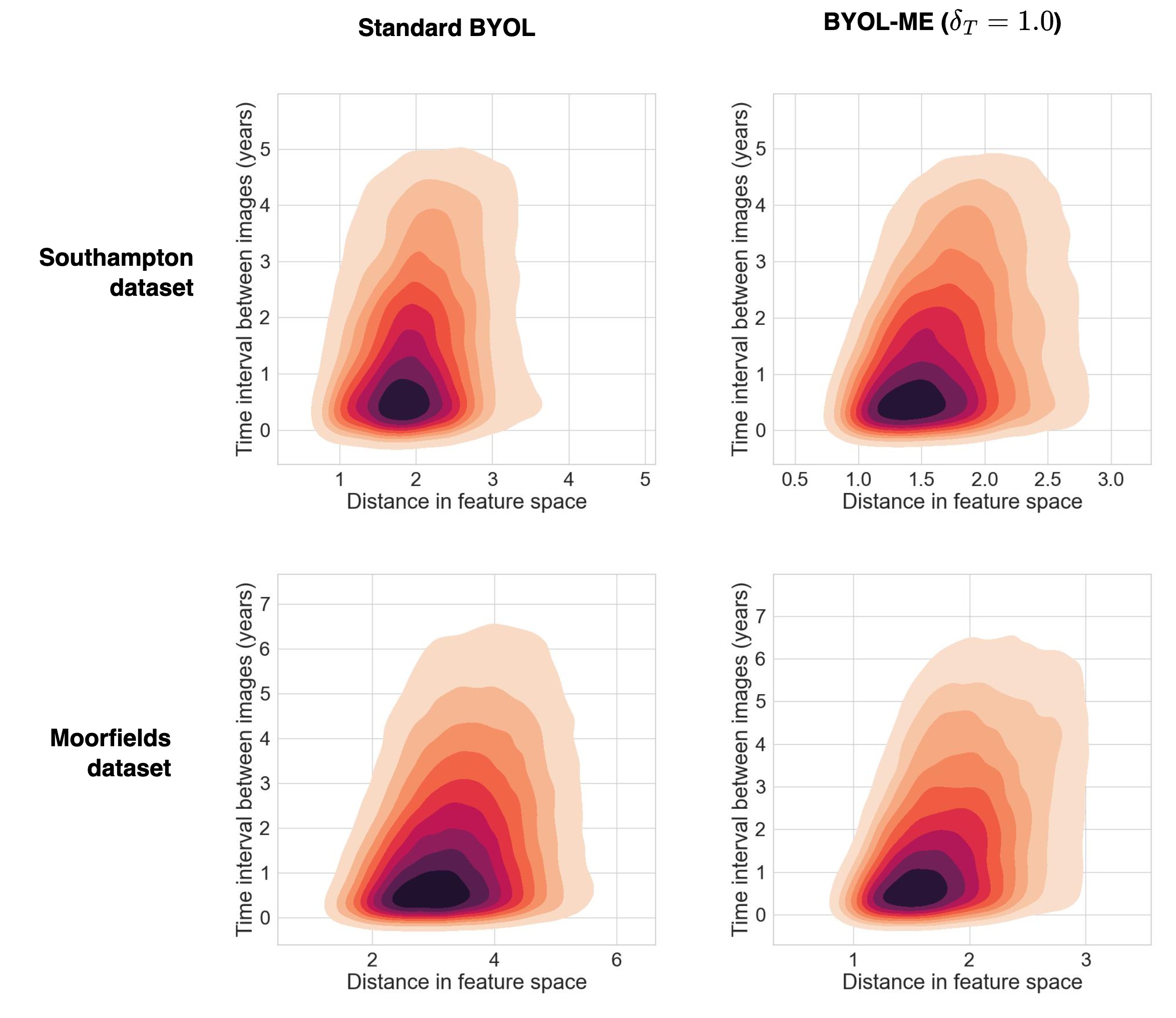}
\caption{{Distributions of times between longitudinal images of the same eye against distance measured in the feature space on unseen test data. Our metadata-enhanced learning explicitly encourages that images of the same eye should, on a small time interval, show similar disease-related features and be separated by a smaller distance in the feature space. Emergence of this relationship can already be seen for standard contrastive learning.}}
\label{fig:dist_vs_dt}
\end{figure*}

{
Our metadata pairings defined in Equation \ref{eq:S} permit only positive relationships between images from the same longitude. This introduces a theoretical degenerate solution in which images from the same eye are mapped to the same latent feature point. In analogy to BYOL \citep{grill2020bootstrap} we find that this degenerate solution reliably does not occur (see Figure \ref{fig:dist_vs_dt}). Moreover, as expected relationships between embeddings of images of the same eye over time respect clinical expectations. Namely, embeddings of images that were taken further apart in time, over which more progression can occur, are typically more distant. While this trend holds in most cases, it highlights two exceptions. Firstly, some images taken over a small window are in fact distant in feature space. This reflects that AMD can progress very rapidly in a short space of time, such as conversion to the late stage. Conversely, some patients progress very slowly and travel a relatively short distance in feature space even over a long time period.\\
In a second robustness test, metadata-enhanced learning did not improve performance on predicting the time of acquisition, a task unrelated to disease state (see Table \ref{tab:timeofday}).
}

\begin{figure*}[t]
    \centering
    \subfloat {\includegraphics[width=0.95\linewidth]{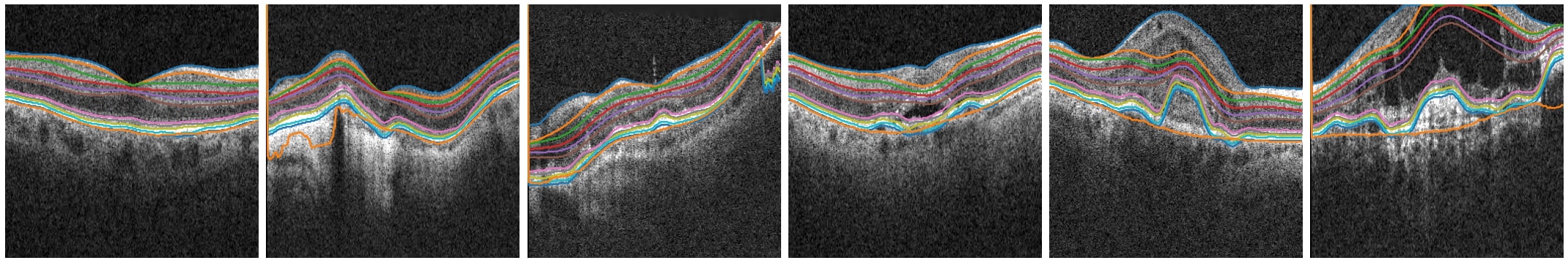}
    }
\caption{{Performance of the IOWA tool on example OCT images from the Southampton dataset with no pathology, hypertransmission, subretinal fluid and intraretinal fluid. In most cases it is robust to large deformations and presence of fluid.}}
\end{figure*}

\begin{figure*}[t]
    \centering
    \includegraphics[width=0.95\linewidth]{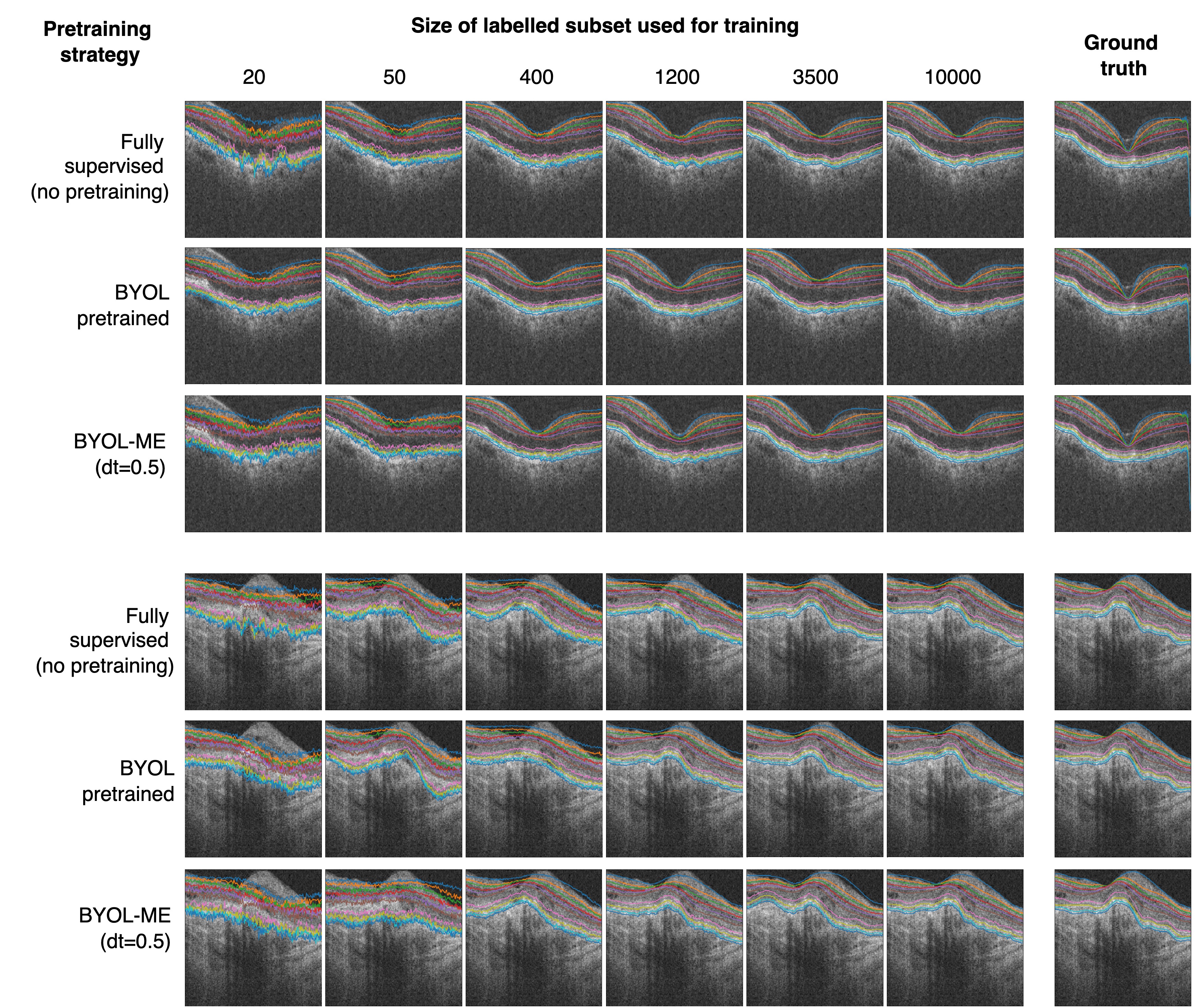}
    \caption{Performance on downstream segmentation of retinal layers using no pretraining (fully supervised), pretraining with BYOL and with BYOL-ME($\delta_T = 0.5$) finetuned using varying sizes of subsets of the labelled data.}
    \label{fig:layer_segmentations}
\end{figure*}

\begin{figure*}[t]
    \centering
    \includegraphics[width=0.85\linewidth]{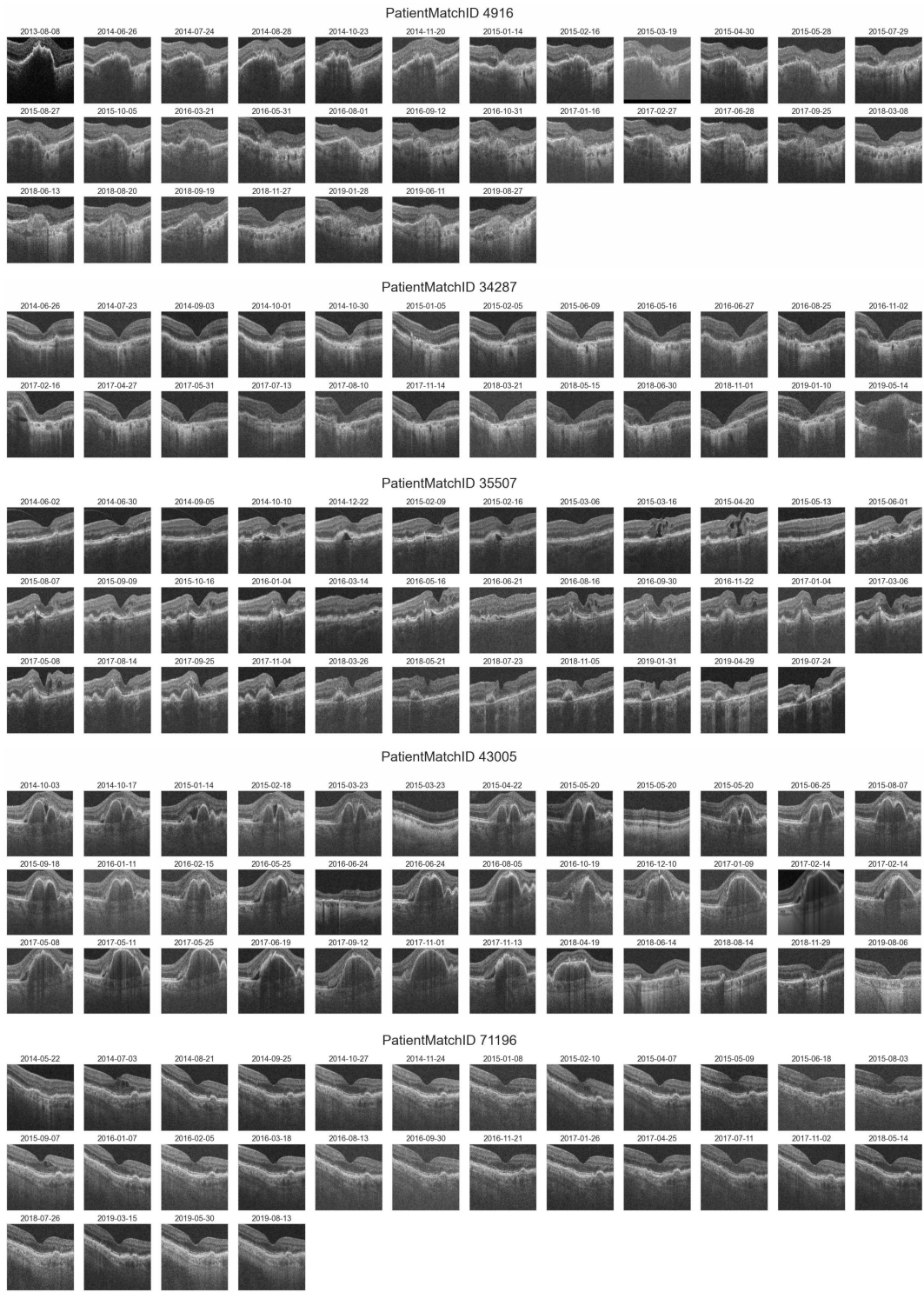}
    \caption{Longitudinal scanning series examples from the Southampton dataset. Most feature growing hypertransmission (seen as signal below the retina) which is evidence for growing atrophy of the photoreceptors.}
    \label{fig:longitude_examples}
\end{figure*}

\begin{table}[]
\centering
\caption{{MAE in hours on test set of models on predicting the time of day of image acquisition. Models were finetuned on 10,000 images from the Southampton dataset.}}
\label{tab:timeofday}
\begin{tabular}{r|cc|cc}
Model & SimCLR & BYOL & \begin{tabular}[c]{@{}c@{}}SimCLR-ME \\ ($\delta_T$=1.0)\end{tabular} & \begin{tabular}[c]{@{}c@{}}BYOL-ME \\ ($\delta_T$=1.0)\end{tabular} \\ \hline
MAE   &     2.65     &  2.65    & 2.65&    2.66  
\end{tabular}
\end{table}

\end{document}